\documentclass[runningheads]{llncs}
\usepackage[T1]{fontenc}
\usepackage{graphicx}
\usepackage{booktabs}
\usepackage[misc]{ifsym}

\usepackage{amsmath}
\usepackage{amssymb}
\usepackage{amsfonts}
\usepackage{color}
\usepackage{bm}
\usepackage{bbm}
\usepackage{multirow}
\usepackage[normalem]{ulem}
\useunder{\uline}{\ul}{}

\usepackage{tabularx}
\usepackage{algorithm}
\usepackage{algorithmic}
\usepackage{tipa}
\usepackage{wrapfig}
\usepackage{float}
\usepackage{enumitem}
\usepackage{bbm}
\usepackage{rotating}
\usepackage{url}

\begin{document}

\title{XicorAttention: Time Series Transformer Using Attention with Nonlinear Correlation}

\titlerunning{XicorAttention: Attention with Nonlinear Correlation}

\author{Daichi Kimura\inst{1, 2} \and
Tomonori Izumitani\inst{1} \and
Hisashi Kashima\inst{2}
}

\authorrunning{D. Kimura et al.}

\institute{
  NTT Communications, Japan \email{\{d.kimura,tomonori.izumitani\}@ntt.com}
\and
Kyoto University, Japan \email{kashima@i.kyoto-u.ac.jp}
}

\maketitle

\begin{abstract} %
Various Transformer-based models have been proposed for time series forecasting. 
These models leverage the self-attention mechanism to capture long-term temporal or variate dependencies in sequences. 
Existing methods can be divided into two approaches: (1) reducing computational cost of attention by making the calculations sparse, and (2) reshaping the input data to aggregate temporal features. 
However, existing attention mechanisms may not adequately capture inherent nonlinear dependencies present in time series data, leaving room for improvement. 
In this study, we propose a novel attention mechanism based on Chatterjee's rank correlation coefficient, which measures nonlinear dependencies between variables. 
Specifically, we replace the matrix multiplication in standard attention mechanisms with this rank coefficient to measure the query-key relationship. 
Since computing Chatterjee's correlation coefficient involves sorting and ranking operations, we introduce a differentiable approximation employing SoftSort and SoftRank. 
Our proposed mechanism, ``XicorAttention,'' integrates it into several state-of-the-art Transformer models. 
Experimental results on real-world datasets demonstrate that incorporating nonlinear correlation into the attention improves forecasting accuracy by up to approximately 9.1\% compared to existing models.

\keywords{Differentiable sorting and ranking \and Chatterjee's rank correlation coefficient \and Accuracy Improvement.}
\end{abstract}

\section{Introduction}
Multivariate time series forecasting is a fundamental task in time series data analysis, especially in areas, such as sensor data from machinery \cite{zhou2021informer}, financial market transaction histories \cite{lai2018modeling}, and weather forecasting \cite{weatherdataset}. 
Inspired by the success of Transformer models \cite{vaswani2017attention} in natural language processing and computer vision, numerous Transformer-based forecasting models have been proposed for time series analysis. 
The Transformer architecture stacks self-attention layers, thereby allowing it to capture long-term dependencies in sequential data.
Existing Transformer-based methods typically adopt one of two main approaches to capture long-term dependencies. 
The first approach reduces the computational cost of attention, enabling the processing of longer input sequences. 
Informer \cite{zhou2021informer} uses a KL-divergence-based metric to select only the top-$k$ queries containing important information, thereby efficiently reducing computational complexity. 
Autoformer \cite{wu2021autoformer} introduces a self-correlation mechanism to aggregate similar sequences along the temporal dimension, considerably lowering computational cost. 

The second approach reshapes or segments the input data to reduce token length and enhance computational efficiency. 
For example, PatchTST \cite{Nie2022ATS} splits the input sequence into shorter patches, significantly reducing the input token length. 
Consider a univariate time series $\bm{x}\in\mathbb{R}^T$, patching with size $P$ and stride $S$ results in a transformed series $\mathbf{X}_{\mathrm{patch}}\in\mathbb{R}^{N\times P}$, where $N=\lfloor \frac{T-P}{S} \rfloor + 2$. 
Each token (patch) retains local temporal information, enabling comprehensive semantic representation. 
Similarly, iTransformer \cite{liu2023itransformer} explicitly captures correlations between variables by transposing patched series before feeding them into the Transformer. 
Recent methods, such as Crossformer \cite{zhang2023crossformer} and TimeXer \cite{wang2024timexer}, combine these approaches to simultaneously model temporal and inter-variable dependencies. 
Most existing methods emphasize on reducing attention’s computational cost or transforming the input data structure, rather than directly enhancing the attention mechanism itself.

However, time series data inherently exhibit nonlinear dynamics and complex interactions among variables \cite{sugihara1990nonlinear,bradley2015nonlinear}. 
Therefore, enhancing the expressive power of the attention mechanism to explicitly model these nonlinearities can substantially improve forecasting accuracy. 
In this study, we propose a novel attention mechanism designed to capture the inherent nonlinear relationships present in multivariate time series. 
Specifically, we incorporate the nonlinear correlation coefficient introduced by Chatterjee \cite{chatterjee2021new} into attention calculations. 

In our experiments on benchmark datasets for multivariate long-term forecasting, we replaced the attention layers of state-of-the-art Transformer-based models with the proposed nonlinear attention mechanism. 
Experimental results demonstrate that our approach effectively enhances the forecasting accuracy of Transformer models. Our contributions are as follows:
\begin{itemize} 
  \item We propose XicorAttention, a novel attention mechanism leveraging Chatterjee's correlation coefficient $\xi$ to effectively capture nonlinear dependencies inherent in time series data.
  \item Since computing Chatterjee's $\xi$ correlation coefficient involves sorting and ranking operations, it is not directly differentiable. To overcome this, we employ differentiable approximations using SoftSort and SoftRank, enabling easy integration into existing Transformer models by replacing the original attention layers.
  \item Extensive experiments demonstrate that Transformer models incorporating XicorAttention achieve superior forecasting accuracy compared to baseline models, with up to approximately 9.1\% improvement in performance.
\end{itemize}

\section{Formulation and Related Work}

\subsection{Problem Formulation}
Let $\mathbf{X}_{1:T} = \{\bm{x}_1, \ldots, \bm{x}_{T}\} \in \mathbb{R}^{T\times C}$ be historical observations of multivariate time series, where $T$ is the lookback window length and $C$ is the number of variables. 
The multivariate forecasting task aims to predict the next $H$-steps forecasting horizon, $\mathbf{X}_{T+1:T+H} = \{\bm{x}_{T+1}, \ldots, \bm{x}_{T+H}\} \in \mathbb{R}^{H\times C}$. 
Consider historical data $\mathbf{X}_{1:T}$, the forecasting model generates predictions as $\hat{\mathbf{X}}_{T+1:T+H} = f_w(\mathbf{X}_{1:T})$, where $f_w(\cdot)$ denotes the forecasting model parameterized by $w$.
The objective of model parameter optimization is to minimize the prediction error between the ground truth $\mathbf{X}_{T+1:T+H}$ and its prediction $\hat{\mathbf{X}}_{T+1:T+H}$.

\subsection{Na\"{\i}ve Attention}
Consider query and key-value pairs, attention calculates the similarity between the query and the key, and then employs this similarity score as a weight to aggregate the corresponding values. 
This effectively extracts important information from the input data. 
Scaled dot-product attention (hereafter simply referred to as attention), proposed by Vaswani et al. \cite{vaswani2017attention}, computes the attention score from the dot product of query and key vectors, and then applies a softmax function to obtain the attention weights. 
In self-attention, the same input data is used to derive queries, keys, and values. 
Consider an input $\mathbf{X}_{1:T} \in \mathbb{R}^{T\times C}$, we apply affine transformations using learnable parameters $\mathbf{W}_Q, \mathbf{W}_K, \mathbf{W}_V \in \mathbb{R}^{C \times D}$ to compute the queries, keys, and values, respectively, as follows:
\begin{align}
  \mathbf{Q}=\mathbf{X}\mathbf{W}_Q,\quad
  \mathbf{K}=\mathbf{X}\mathbf{W}_K,\quad
  \mathbf{V}=\mathbf{X}\mathbf{W}_V \in \mathbb{R}^{T\times D},
\end{align}
where $D$ denotes the dimension of the attention, commonly called the model dimension.
When employing multi-head attention, these queries, keys, and values are further split into $n_{\text{head}}$ separate attention heads. Each attention head thus has a reduced dimension $d=D/n_{\text{head}}$, allowing the model to jointly capture information from multiple representation subspaces. 
The attention calculation within each head is as follows: 
\begin{align}
A(\mathbf{Q,K,V})=\mathrm{softmax} \left( \frac{\mathbf{QK}^\mathrm{T}}{\sqrt{ d }}\right) \mathbf{V}. 
\end{align}
Focusing on the attention scores $\mathbf{QK}^\mathrm{T}$, the dot product between the $i$-th row vector $\bm{q}_i\in\mathbb{R}^d$ of $\mathbf{Q}$ and the $j$-th column vector $\bm{k}_j^\mathrm{T}\in\mathbb{R}^d$ of $\mathbf{K}^\mathrm{T}$ can be interpreted as measuring the cross-correlation between $\bm{q}$ and $\bm{k}$ \cite{wu2021autoformer,nguyen2025correlated}. 
\begin{align}
  \bm{q}_i\bm{k}_j^\mathrm{T} = \sum_{l}^d q_{i,l}k_{j,l} \sim \rho(\bm{q}_i,\bm{k}_j) .
\end{align}
Pearson's correlation coefficient $\rho$ measures the linear relationship between two variables $(\bm{x},\bm{y}):=(x_1,y_1),\ldots,(x_n,y_n)$. If both variables are mean-centered, 
\begin{align}
  \rho(\boldsymbol{x},\boldsymbol{y})=\frac{\mathrm{Cov}(\bm{x}, \bm{y})}{\sqrt{\sigma_x^2}\sqrt{\sigma_y^2}} = \frac{\sum_{i=1}^n x_i y_i}{\sqrt{\sum_{i=1}^n x_i^2}\sqrt{\sum_{i=1}^n y_i^2}},
\end{align}
where $\mathrm{Cov}$ is the covariance, and $\sigma_x^2, \sigma_y^2$ are the variances. 
The coefficient $\rho$ ranges from $-1$ to $1$, with values near 1 indicating strong positive linear correlation, values close to -1 indicating strong negative linear correlation, and values close to 0 indicating no correlation. 

However, Pearson's correlation cannot accurately detect nonlinear relationships, resulting in correlation values close to zero even when variables are nonlinearly related. 
Therefore, applying standard attention mechanisms to time series data with inherent nonlinear relationships may cause an inadequate representation of important temporal dynamics and inter-variable dependencies. 

\subsection{Transformer-based Models for Time Series Forecasting}
Various Transformer-based models have been proposed to adapt the original Transformer architecture \cite{vaswani2017attention} for time series forecasting tasks.
Specifically, most existing methods primarily focus on reducing the quadratic computational complexity of attention mechanisms, either by introducing approximations in attention or reshaping input data.

\subsubsection{Improving Computational Efficiency of Attention.}
Several methods have been proposed to improve computational efficiency by modifying the standard attention mechanism. 
LogSparse attention \cite{li2019enhancing} reduces computational cost by selecting time steps at exponential intervals. 
Informer \cite{zhou2021informer} computes attention scores only for the top-$u$ dominant queries identified via KL-divergence. 
Autoformer \cite{wu2021autoformer} treats attention as an autocorrelation operation, efficiently aggregating temporal dependencies while reducing computational complexity. 
FEDformer \cite{zhou2022fedformer} maps the attention calculation into the frequency domain using the Fourier transform and randomly selects frequency components to reduce computational complexity. 

\subsubsection{Reshaping Input Data to Improve Efficiency.}
In standard Transformer and the previously mentioned models, each time point $t$ in a multivariate time series $\mathbf{X}_{1:T}$ is represented as one input token $\bm{x}_t \in \mathbb{R}^{1 \times C}$. However, this representation causes significant computational overhead, motivating recent Transformer-based methods to introduce various techniques for reshaping or segmenting the input data. 

PatchTST \cite{Nie2022ATS} decomposes a multivariate time series $\mathbf{X} \in \mathbb{R}^{T\times C}$ into individual univariate time series $\bm{x} \in \mathbb{R}^{T\times 1}$ and then splits each univariate series into smaller subseries (patches). 
Consider a patch length $P$ and stride $S$, the resulting input to the Transformer becomes $\mathbf{X}_{\mathrm{patch}}\in{\mathbb{R}^{N \times P}}$, where $N=\lfloor \frac{T-P}{S} \rfloor + 2$. This patching technique efficiently captures local subsequence information and global temporal dependencies while significantly reducing computational complexity. Finally, the predictions from each univariate series are aggregated to produce a multivariate forecast. 
iTransformer \cite{liu2023itransformer} explicitly captures dependencies among variables by transposing the input matrix, resulting in a Transformer input of shape $\mathbf{X}^\mathrm{T} \in \mathbb{R}^{C \times T}$.
TimeXer \cite{wang2024timexer} aims to capture both temporal and inter-variable dependencies by combining patching with self-attention and cross-attention mechanisms. 
Notably, these models utilize the standard attention mechanism without additional modifications to the attention calculation itself.

\section{Proposed method: XicorAttention}
\begin{figure}[t] 
  \centering
  \includegraphics[width=0.95\textwidth]{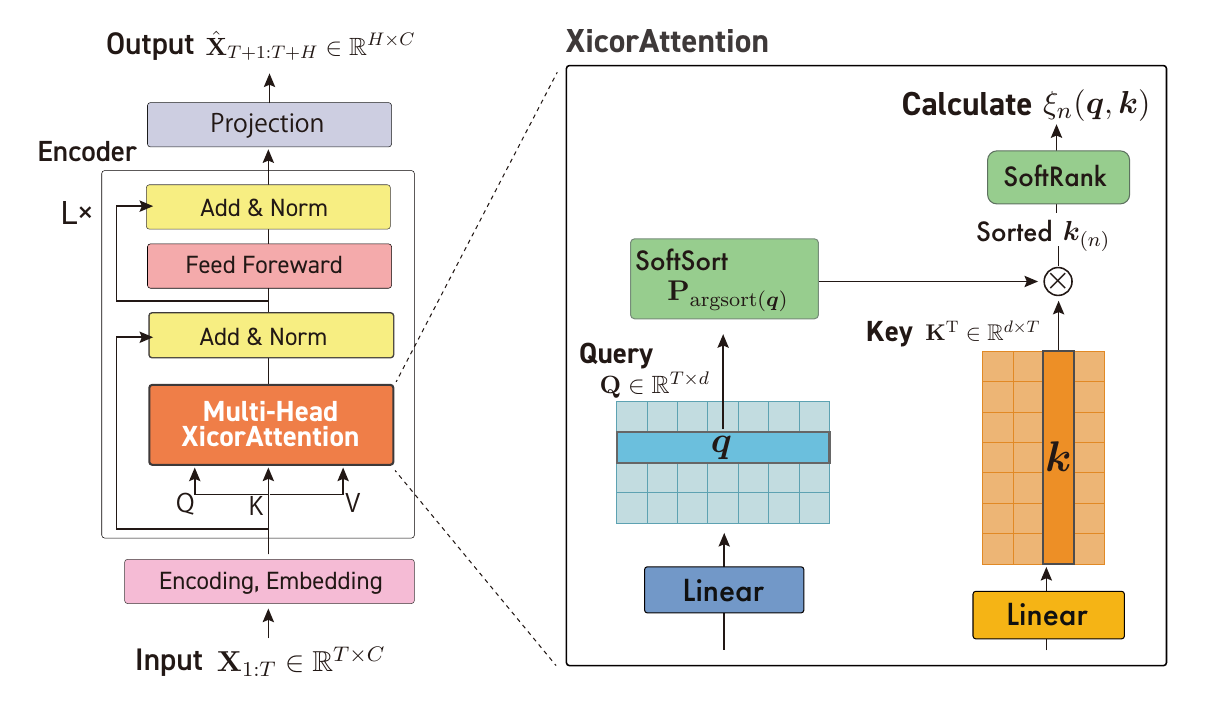}
  \caption{Overview of the proposed XicorAttention architecture integrated into an encoder-based Transformer forecasting model (left). 
  In XicorAttention (right), attention weights are computed based on Chatterjee's correlation coefficient $\xi_n(\bm{q}, \bm{k})$ between queries and keys. Since $\xi_n$ involves non-differentiable sorting and ranking operations, we introduce differentiable approximations using SoftSort and SoftRank to enable gradient-based parameter updates. For simplicity, the value computation is omitted in this illustration.
  } 
  \label{fig:arch}
\end{figure}

We propose a novel attention mechanism named \textit{XicorAttention} \footnote{The pronunciation is \textipa{[saIkO: - \textipa{2}\textipa{"}ten\textipa{S}\textipa{2}n]}}, illustrated in Figure~\ref{fig:arch}, to capture nonlinear dependencies in time series data. 
Contrarily to traditional attention mechanisms that rely on linear interactions through matrix products, our method employs Chatterjee's rank-based nonlinear correlation coefficient $\xi$ \cite{chatterjee2021new}, enabling effective modeling of nonlinear relationships. 
As discussed in Section~\ref{sec:xi}, the coefficient $\xi_n$ naturally converges almost to a limit within the range $[0,1]$ as $n \to \infty$. Specifically, $\xi_n=0$ indicates independence between variables, while $\xi_n=1$ corresponds to a deterministic measurable relationship. 

In the following subsection, we first introduce Chatterjee's $\xi$ correlation coefficient, the core concept behind our method, and then describe our approach for differentiable. 
Note that we use $n$ (the sample size for $\xi_n$) and $d$ (the dimension of each attention head) interchangeably for notational simplicity. 

\begin{algorithm}[t]
  \caption{Chatterjee's $\xi$ correlation coefficient}
  \begin{algorithmic}[1]
  \REQUIRE $(X, Y):=(X_{1}, Y_{1}),\ldots, (X_{n}, Y_{n})$
  \STATE Sort $X$ in ascending order
        \\ $(X_{(1)}, Y_{(1)}), \ldots (X_{(n)}, Y_{(n)})$
  \STATE Rank $Y_{(i)}$ in ascending order
        \\ $r_{1}, \ldots, r_{n}$
  \STATE Calculate $\xi_n$
        \\ \renewcommand{\arraystretch}{4}$\xi_{n}(X, Y) := 1- \frac{3 \sum_{i=1}^{n-1} |r_{i+1}-r_{i}|}{n^2-1}$
  \end{algorithmic}
  \label{alg:xi}
\end{algorithm}

\subsection{Chatterjee's $\xi$ correlation coefficient}\label{sec:xi}
Chatterjee's $\xi$ correlation coefficient (hereafter simply referred to as $\xi$) \cite{chatterjee2021new} is a rank-based measure, similar to Spearman's rank correlation, designed to detect nonlinear relationships between two variables $X$ and $Y$. The coefficient $\xi$ has the following desirable properties:
\begin{itemize}
  \item It is as easy to compute as classical correlation coefficients, such as Pearson's correlation or Spearman's correlation.
  \item It assumes values in the interval $[0, 1]$ for large samples, with clear interpretations: $\xi=0$ indicates independence between $X$ and $Y$, $\xi=1$ indicates that $Y$ is a measurable function of $X$ (i.e., $Y=f(X)$) 
  \item It requires no assumptions about the distribution or functional forms relating $X$ and $Y$. 
  \item It is robust and effective in capturing oscillatory or highly nonlinear dependence patterns. 
\end{itemize}
The outline of the calculation is shown in Algorithm~\ref{alg:xi}. 
This coefficient is computed as follows. 
Consider independent samples $(X_{1}, Y_{1}),\ldots (X_{n}, Y_{n})$ (where $Y$ is not a constant), assume for simplicity that there are no ties among $X_i$ or $Y_i$. 
First, sort the pairs in ascending order based on $X_i$, yielding $(X_{(1)}, Y_{(1)}), \ldots (X_{(n)}, Y_{(n)})$ with $X_{(1)} < X_{(2)} < \ldots < X_{(n)}$. 
Subsequently, assign ranks $r_i$ to $Y_{(i)}$ in ascending order, i.e. the smallest $Y_{(i)}$ has rank $1$, and the largest has rank $n$. 
Therefore, the $\xi_n(X, Y)$ coefficient is calculated by: 
\begin{align}
\xi_{n}(X, Y) := 1- \frac{3 \sum_{i=1}^{n-1} |r_{i+1}-r_{i}|}{n^2-1}.
\end{align}
The intuition behind this formula is straightforward. If there is a strong dependence between $X$ and $Y$, the ranks of $Y$ will change smoothly when data is sorted by increasing values of $X$, resulting in small differences $|r_{i+1}-r_{i}|$; thus, $\xi_n(X,Y)$ will be close to $1$. 
Otherwise, the ranks $Y$ appear randomly with respect to the ordered $X$, causing larger rank differences and thus $\xi_n(X,Y)$ near $0$.

Chatterjee showed that as the sample size $n\rightarrow \infty$, the coefficient $\xi_n(X,Y)$ converges almost to a deterministic limit $\xi(X,Y)$, defined as follows: 
\begin{align}
  \xi(X, Y) := \frac{\int \mathrm{Var}\left(\mathbb{E}[\mathbbm{1}_{Y\ge t}|X]\right)d\mu(t)}{\int\mathrm{Var}\left(\mathbbm{1}_{Y\ge t}\right) d\mu(t)},
\end{align}
where $\mu$ denotes the probability distribution of $Y$. Dette et al. \cite{dette2013copula} introduced similar copula-based estimator for this limiting quantity. 
The intuitive interpretation of $\xi(X, Y)$ is describe as follows. 
The numerator measures how effectively $X$ explains the distribution of $Y$ at various thresholds $t$. 
Specifically, $\mathrm{Var}(\mathbb{E}[\mathbbm{1}_{Y\ge t}|X])$ is the conditional probability that $Y$ exceeds $t$ given $X$. 
If $X$ strongly explains $Y$, this conditional expectation varies substantially with changes in $X$, resulting in large variance. The denominator normalizes this measure by the inherent variability of $Y$ itself, without considering $X$. 
Thus, $\xi(X,Y)=0$ if and only if $X$ and $Y$ are independent, and $\xi=1$ if and only if $Y$ is a deterministic measurable function of $X$.

\subsection{Differentiable $\xi_n$ coefficient}
In conventional attention, the relevance between query and key is measured by their matrix product. 
However, the proposed method employs $\xi_n$ to capture nonlinear relationships. 
Specifically, the attention weights between query vector $\bm{q}$ and key vector $\bm{k}$ are computed as $\xi_{n}(\bm{q}, \bm{k})$. 

However, calculating $\xi_{n}(\bm{q}, \bm{k})$ involves sorting and ranking operations (Algorithm~\ref{alg:xi}-1,2), which are not explicitly differentiable with respect to the model parameters $\mathbf{W}_Q$ and $\mathbf{W}_K$. 
Therefore, we employ differentiable sorting and ranking techniques, specifically SoftSort \cite{prillo2020softsort} and FastSoftRank \cite{pmlr-v119-blondel20a}, allowing the gradients to propagate effectively during training. 

A detailed explanation of SoftSort and FastSoftRank is provided in the following section; however, we have briefly summarized their computational costs here. 
The computational costs of SoftSort and FastSoftRank are $\mathcal{O}(n^2)$ and $\mathcal{O}(n\log n)$, respectively. Thus, the overall computational complexity of the proposed method is $\mathcal{O}(n^2)$.

\subsection{SoftSort}
To calculate $\xi_n$, it is necessary to sort the pairs $\bm{q}$ and $\bm{k}$ (Algorithm~\ref{alg:xi}-1). 
Specifically, we sort $\bm{q}$ in ascending order and rearrange $\bm{k}$ in the same order. 
To archive this sorting operation, we introduce a permutation matrix $\mathbf{P}_{\pi} \in \mathbb{R}^{n\times n}$. 
Consider a permutation $\pi:\{1,\ldots,n\}\rightarrow \{1,\ldots, n\}$, the corresponding permutation matrix $\mathbf{P}_{\pi} \in \mathbb{R}^{n\times n}$ is a binary matrix defined as:
\begin{align}
  \mathbf{P}_{\pi}[i,j] = 
  \begin{cases}
    1 & \mathrm{if} \quad j = \pi_i \\
    0 & \mathrm{otherwise}.
  \end{cases}
  \label{eq:permmat}
\end{align}
The permutation returned by the $\texttt{argsort}(\bm{q})$ operator, which sorts the vector $\bm{q}$ in descending order, can be represented as a permutation matrix $\mathbf{P}_{\texttt{argsort}(\bm{q})}$. For sorting in ascending order, we reverse the sign of $\bm{q}$.
For example, consider $\bm{q}=[1.2, 9.3, 1.7, 3.6]^\mathrm{T}$, the permutation returned by $\texttt{argsort}(\bm{q})$ is $[2,4,3,1]$, thus, the sorted vector $\bm{q}_{(n)}$ can be computed as:
\begin{align}
  \bm{q}_{(n)}= \mathbf{P}_{\texttt{argsort}(\bm{q})} \bm{q}.
\end{align}
Similarly, to obtain the corresponding vector $\bm{k}_{(n)}$, sorted according to the ascending order of $\bm{q}$, we compute
\begin{align}
  \bm{k}_{(n)} =\mathbf{P}_{\texttt{argsort}(-\bm{q})} \bm{k} .
\end{align}

Because $\mathbf{P}_{\texttt{argsort}}$ is discrete, it has zero gradient almost everywhere, preventing direct gradient-based optimization. 
To overcome this limitation, several continuous relaxation methods for the permutation matrix have been proposed \cite{grover2018stochastic,mena2018learning,prillo2020softsort}. 
In this work, we adopt the simplest of these, SoftSort \cite{prillo2020softsort}, defined as:
\begin{align}
  \hat{\mathbf{P}}_{\texttt{argsort}(\bm{q})}:=\mathrm{SoftSort}_\tau(\bm{q}) = \mathrm{softmax}\left( \frac{ -d(\mathrm{sort}(\bm{q}) \mathbf{1}^\mathrm{T} , \mathbf{1}\bm{q}^\mathrm{T})}{\tau}\right),
  \label{eq:softsort}
\end{align}
where $\mathbf{1}=[1,\ldots,1]^\mathrm{T} \in \mathbb{R}^n$ is the all-one vector, the softmax function is applied row-wise, $d(x,y)$ is a differentiable distance function (e.g. L1 norm $|x-y|$), $\tau$ is a temperature parameter controlling the sharpness of the approximation. 
Intuitively, the numerator $-d (\mathrm{sort}(\bm{q}) \mathbf{1}^\mathrm{T} , \mathbf{1}\bm{q}^\mathrm{T})$ how closely each original element aligns with each sorted position, acting as similarity measure. 
Larger similarities (smaller distances) receive high probabilities through the softmax operation, enabling a differentiable approximation of the sorting operation. 
Although SoftSort effectively relaxes the discrete permutation, it has a computational complexity of $\mathcal{O}(n^2)$.

In practice, since SoftSort produces a soft permutation matrix whose rows can be interpreted as similarities (or soft assignments) of each original element ot sorted positions, the resulting sorted vector $\hat{\mathbf{P}}_{\texttt{argsort}(\bm{q})}\bm{k}$ may differ from the exact sorted order. Particularly, when elements of $\bm{q}$ have similar values, the soft permutation matrix spreads assignment weight across multiple positions. 
To address this issue, we apply a straightforward-through trick, enforcing an exact permutation during the forward pass without affecting gradient computation, as follows:
\begin{align}
\hat{\mathbf{P}}_{\texttt{argsort}}^{\mathrm{ST}}{(\bm{q})} = \hat{\mathbf{P}}_{\texttt{argsort}(\bm{q})} + \mathrm{sg} \left( \mathrm{onehot\text{-}argmax} (\hat{\mathbf{P}}_{\texttt{argsort}(\bm{q})}) - \hat{\mathbf{P}}_{\texttt{argsort}(\bm{q})} \right),
\end{align}
where $\mathrm{sg}(\cdot)$ is the stop-gradient operation and $\mathrm{onehot\text{-}argmax}(\mathbf{P})$ is a row-wise operation converting each row of $\mathbf{P}$ to a one-hot vector, where the position of the maximum value is set to 1 and all other entries to 0.

\subsection{FastSoftRank}
To compute Algorithm~\ref{alg:xi}-2, as discussed in the previous section, a continuous relaxation of the ranking operation is required. It is well known that the rank operator $\mathbf{P}_{\texttt{rank}}$ is defined as $\mathbf{P}_{\texttt{rank}}=\mathbf{P}_{\texttt{argsort}}^\mathrm{T}$ \cite{prillo2020softsort}. While the SoftSort method can be directly applied, it has $\mathcal{O}(n^2)$ complexity. 

Therefore, we employ a faster method, FastSoftRank,which is proposed by Blondel et al. \cite{pmlr-v119-blondel20a}. 
FastSoftRank is a differentiable ranking operator that approximates standard ranking by projecting onto the permutahedron, the convex hull of all permutations. 
This approach enables smooth rank approximations, making it suitable for integration into gradient-based optimization frameworks. 

Here, we explain FastSoftRank in a concrete formulation, rather than the general form presented in previous study \cite{pmlr-v119-blondel20a}. 
A permutahedron is a convex polytope defined as the convex hull of all permutations of a particular vector. 
Consider a fixed vector $\bm{\rho}\in\mathbb{R}^n$, the permutahedron $\mathcal{P}(\bm{\rho})$ is defined as:
\begin{align}
\mathcal{P}(\bm{\rho}):= \operatorname*{conv}({\bm{\rho}_\sigma:\sigma\in\Sigma}) \subset \mathbb{R}^n,
\end{align}
where $\operatorname*{conv}$ is the convex hull, $\bm{\rho}_\sigma$ is a vector obtained by the elements of $\bm{\rho}$ according to a permutation $\sigma$, and $\Sigma$ is the set of all permutations of size $n$. 
Each vertex of this polytope corresponds to one particular permutation of the the elements in $\bm{\rho}$. 
For instance, for $\bm{\rho}=(1,2,3)$, the permutahedron $\mathcal{P}(\bm{\rho})$ has vertices corresponding to all $3!=6$ permutations, i.e., $(1,2,3), (1,3,2), \ldots, (3, 2, 1)$. 
The standard (non-differentiable) ranking operation to $\bm{k}$ can be formulated as a projection onto this permutahedron:
\begin{align}
  r(\bm{k}) = \operatorname*{arg\,max}_{\bm{y}\in\mathcal{P}(\bm{\rho})} \left<\bm{y}, -\bm{k}\right>,
\end{align}
where, $\left<\bm{y}, -\bm{k}\right>$ is inner product of vectors. 
Specifically, the ranking operation finds the vertex $\bm{y}$ of the permutahedron $\mathcal{P}(\bm{\rho})$ that has the largest inner product with the $-\bm{k}$. 
This original problem is non-differentiable owing to its discrete nature. 
Therefore, Blondel et al. \cite{pmlr-v119-blondel20a} introduced a strongly convex regularization term, such as the quadratic regularization $\Phi(\bm{y})=\frac{1}{2}\lVert \bm{y} \rVert^2$, resulting in a smooth and differentiable approximation of the rank operation: 
\begin{align}
  r_\Phi^\varepsilon(\bm{k}) &= \operatorname*{arg\,max}_{\bm{y}\in\mathcal{P}(\bm{\rho})} \left<\bm{y}, -\bm{k}\right> - \varepsilon \Phi(\bm{y})  =\operatorname*{arg\,min}_{y\in\mathcal{P}(\bm{\rho})} \frac{1}{2} \left\lVert \bm{y}+\bm{k}/\varepsilon \right\rVert^2, 
\end{align}
where $\varepsilon$ is the regularization strength, as $\varepsilon \rightarrow 0$, soft rank $r_\Phi^\varepsilon(\bm{k})$ converges to original ranking operator. 

This resulting formulation can be reduced to an isotonic optimization problem. Let $s=(-\bm{k}/\varepsilon)_{\sigma(-\bm{k}/\varepsilon)}$ be the vector obtained by sorting $-\bm{k}/\varepsilon$ in descending order. Subsequently, we obtain: 
\begin{align}
  r_\Phi^\varepsilon(\bm{k}) = -\frac{\bm{k}}{\varepsilon} - \left[ v_\Phi(\bm{s}, \bm{\rho})\right]_{\sigma^{-1}(-\bm{k}/\varepsilon)},
\end{align}
where the function $v_\Phi$ is known as isotonic regression:
\begin{align}
  v_\Phi(\bm{s}, \bm{\rho}) = \operatorname*{arg\,min}_{v_1\ge\ldots\ge v_n} \frac{1}{2} \left\lVert \bm{v}-(\bm{s}-\bm{\rho}) \right\rVert^2.
\end{align} 
Intuitively, the vector $\bm{v}_\Phi$ represents a monotonic correction applied to the particular vector $-\bm{k}/\varepsilon$ to bring it closer to the ideal descent ranking vector $\bm{\rho}=(n, n-1, \ldots, 1)$. 
Because the resulting solution $\bm{v}_\Phi$ is naturally in sorted order, we employ the inverse permutation $\sigma^{-1}(-\bm{k}/\varepsilon)$ to return it back to the original ordering of $\bm{k}$. 

The isotonic regression problem can be efficiently solved using the Pool Adjacent Violators (PAV) algorithm \cite{best2000minimizing}, which has $\mathcal{O}(n)$ complexity. 
Consider that sorting the original vector requires $\mathcal{O}(n\log n)$ complexity, the total complexity is $\mathcal{O}(n\log n)$ , and its gradient can be computed in $\mathcal{O}(n)$. 

\section{Experiments}
Here, we evaluate the proposed XicorAttention on multivariate long-term time series forecasting problems. 
In our experiments, we integrated XicorAttention into three models: PatchTST~\cite{Nie2022ATS}, iTransformer~\cite{liu2023itransformer}, and TimeXer~\cite{wang2024timexer}. 

\begin{table}[t]
  \centering
  \caption{Overview of the datasets. From the left column, the dataset name, the number of variates, the number of data used for train/validation/test, and the sampling frequency are shown.}
  \setlength{\tabcolsep}{7pt}
  \begin{tabular}{cccc}
  \hline
  Dataset      & Variates & Size (train, valid, test) & Sampling Freq. \\ \hline
  ETTh1, ETTh2  & 7             & (8545, 2881, 2881)        & Hourly             \\ 
  ETTm1, ETTm2  & 7             & (34465, 11521, 11521)     & 15min              \\ 
  Exchange    & 8             & (5120, 665, 1422)         & Daily              \\ 
  Weather     & 21            & (36792, 5271, 10540)      & 10min              \\ 
  Electricity     & 321            & (18317, 2633, 5261)      & Hourly             \\ 
  Traffic     & 862            & (12185, 1757, 3509)      & Hourly             \\ \hline
  \end{tabular}
  \label{tab:dataset}
\end{table}

\subsection{Datasets}
We conducted extensive experiments on seven real-world datasets \footnote{All datasets are publicly available at \url{https://github.com/thuml/Time-Series-Library}}: ETT (4 subsets), Exchange, Weather, Electricity and Traffic. 
The overview of these datasets is provided in Table~\ref{tab:dataset}. 
Each dataset was split into training, validation, and test sets. The models were trained on the training sets, and their prediction performance was evaluated on the test sets.

\subsection{Baselines}
We selected state-of-the-art Transformer-based models as baselines, including iTransformer \cite{liu2023itransformer}, PatchTST \cite{Nie2022ATS}, FEDformer \cite{zhou2022fedformer}, Informer \cite{zhou2021informer}, Autoformer \cite{wu2021autoformer}, Transformer \cite{vaswani2017attention} were used. 
The implementations were directly taken from the Time-Series-Library \cite{wu2023timesnet}.

\subsection{Implementation}
The lookback window length $T$ was fixed at $T=96$, and we evaluated the models with four forecasting horizons: $H=\{96,192,336,720\}$. 
This forecasting setting has been used in previous studies. 
Our implementation was based on PyTorch\cite{paszke2019pytorch}, and the experiments were conducted on servers with either six NVIDIA RTX A6000 (48 GB) GPUs or eight NVIDIA H100 (80 GB) GPUs. After training, the prediction performance was evaluated on the test sets using a single GPU. 
We used mean absolute error (MAE) and mean squared error (MSE) as evaluation metrics, with smaller values indicating better prediction performance.

\begin{table}[t]
  \centering
  \caption{Experimental results of multivariate long-term time series forecasting. The columns represent the methods and evaluation metrics (MAE, MSE), and the rows correspond to the datasets. The best values are shown in bold, and the second-best values are underlined.}
  \label{tab:result1}
  \resizebox{\textwidth}{!}{
    \begin{tabular}{c|cccccccccccccc}
      \hline
      \multirow{2}{*}{Model} & \multicolumn{2}{c}{Xicor+TimeXer} & \multicolumn{2}{c}{Xicor+iTrans} & \multicolumn{2}{c}{Xicor+Patch} & \multicolumn{2}{c}{FEDformer} & \multicolumn{2}{c}{Informer} & \multicolumn{2}{c}{Autoformer} & \multicolumn{2}{c}{Transformer} \\
                             & \multicolumn{2}{c}{(Ours)}          & \multicolumn{2}{c}{(Ours)}         & \multicolumn{2}{c}{(Ours)}  & \multicolumn{2}{c}{\cite{zhou2022fedformer}}      & \multicolumn{2}{c}{\cite{zhou2021informer}}     & \multicolumn{2}{c}{\cite{wu2021autoformer}}       & \multicolumn{2}{c}{\cite{vaswani2017attention}}        \\ \hline
      Metrics                & MAE             & MSE             & MAE              & MSE           & MAE            & MSE            & MAE       & MSE               & MAE           & MSE          & MAE            & MSE           & MAE            & MSE            \\ \hline
      ETTh1                  & {\ul 0.448}     & 0.461           & 0.455            & 0.466         & \textbf{0.438} & {\ul 0.441}    & 0.458     & \textbf{0.439}    & 0.805         & 1.052        & 0.491          & 0.503         & 0.778          & 0.953          \\
      ETTh2                  & 0.409           & 0.388           & {\ul 0.408}      & {\ul 0.384}   & \textbf{0.402} & \textbf{0.379} & 0.455     & 0.443             & 1.770         & 4.656        & 0.457          & 0.444         & 1.678          & 4.475          \\
      ETTm1                  & \textbf{0.398}  & {\ul 0.386}     & 0.409            & 0.409         & {\ul 0.401}    & \textbf{0.384} & 0.456     & 0.448             & 0.700         & 0.887        & 0.513          & 0.567         & 0.741          & 0.953          \\
      ETTm2                  & \textbf{0.323}  & \textbf{0.276}  & 0.328            & 0.284         & {\ul 0.325}    & {\ul 0.280}    & 0.349     & 0.303             & 0.906         & 1.725        & 0.362          & 0.319         & 0.819          & 1.300          \\
      Exchange               & 0.415           & 0.386           & {\ul 0.407}      & {\ul 0.364}   & \textbf{0.399} & \textbf{0.352} & 0.503     & 0.521             & 1.005         & 1.622        & 0.502          & 0.508         & 0.895          & 1.379          \\
      Weather                & \textbf{0.273}  & \textbf{0.244}  & 0.280            & 0.260         & {\ul 0.277}    & {\ul 0.254}    & 0.361     & 0.312             & 0.551         & 0.629        & 0.414          & 0.390         & 0.574          & 0.643          \\
      Electricity            & {\ul 0.277}     & \textbf{0.179}  & \textbf{0.274}   & {\ul 0.182}   & 0.287          & 0.197          & 0.333     & 0.222             & 0.437         & 0.360        & 0.353          & 0.257         & 0.468          & 0.404          \\
      Traffic                & \textbf{0.292}  & {\ul 0.473}     & 0.388            & 0.570         & {\ul 0.303}    & \textbf{0.472} & 0.379     & 0.610             & 0.480         & 0.850        & 0.412          & 0.662         & 0.366          & 0.668          \\ \hline
      \end{tabular}
  }
\end{table}

\section{Results}
In this section, we comprehensively evaluate the forecasting performance of our XicorAttention. 
First, we compare forecasting models incorporating the proposed XicorAttention with baseline methods that focus on reducing computational cost (Section~\ref{sec:forecasting}). 
Next, we evaluate how much our proposed attention mechanism enhances the forecasting accuracy of existing Transformer-based models utilizing the original self-attention mechanism (Section~\ref{sec:enhance}).

\subsection{Overall Forecasting Performance}\label{sec:forecasting} 
Comprehensive forecasting results across multiple benchmark datasets are summarized in Table~\ref{tab:result1}. %
Results shown here are averaged over four forecasting horizons ($H\in\{96,192,336,720\}$), with the complete results provided in Appendix~\ref{apd:sec:full_performance}. 
The best results are highlighted in bold, and the second-best results are underlined. 
Overall, our proposed XicorAttention consistently outperforms existing state-of-the-art methods on the majority of benchmarks. 

\begin{table}[t]
  \caption{Performance enhancement (\%) by replacing original attention with XicorAttention (+Xicor) in PatchTST, iTransformer, and TimeXer. Bold indicates the better result between original and XicorAttention.}
  \centering
  \begin{tabular}{cccccccc}
    \hline
    \multicolumn{2}{c|}{Models}                                                              & \multicolumn{2}{c}{PatchTST}    & \multicolumn{2}{c}{iTransformer} & \multicolumn{2}{c}{TimeXer}     \\ \hline
    \multicolumn{2}{c|}{Metric}                                                              & MAE            & MSE            & MAE             & MSE            & MAE            & MSE            \\ \hline
    \multicolumn{1}{c|}{\multirow{3}{*}{ETTm2}}       & \multicolumn{1}{c|}{Original}        & 0.331          & 0.280          & 0.330           & 0.279          & 0.319          & \textbf{0.263} \\
    \multicolumn{1}{c|}{}                             & \multicolumn{1}{c|}{\textbf{+Xicor}} & \textbf{0.322} & \textbf{0.268} & \textbf{0.324}  & \textbf{0.271} & \textbf{0.318} & 0.264          \\ \cline{2-8} 
    \multicolumn{1}{c|}{}                             & \multicolumn{1}{c|}{Enhancement}     & 2.91\%         & 4.24\%         & 1.89\%          & 2.68\%         & 0.08\%         & -0.14\%        \\ \hline
    \multicolumn{1}{c|}{\multirow{3}{*}{Exchange}}    & \multicolumn{1}{c|}{Original}        & 0.378          & 0.276          & \textbf{0.369}  & \textbf{0.259} & 0.381          & 0.278          \\
    \multicolumn{1}{c|}{}                             & \multicolumn{1}{c|}{\textbf{+Xicor}} & \textbf{0.361} & \textbf{0.251} & 0.369           & 0.259          & \textbf{0.374} & \textbf{0.270} \\ \cline{2-8} 
    \multicolumn{1}{c|}{}                             & \multicolumn{1}{c|}{Enhancement}     & 4.48\%        & 9.12\%         & -0.10\%         & -0.30\%        & 1.88\%         & 2.96\%         \\ \hline
    \multicolumn{1}{c|}{\multirow{3}{*}{Traffic}}     & \multicolumn{1}{c|}{Original}        & 0.308          & 0.482          & \textbf{0.282}  & \textbf{0.422} & \textbf{0.287} & \textbf{0.465} \\
    \multicolumn{1}{c|}{}                             & \multicolumn{1}{c|}{\textbf{+Xicor}} & \textbf{0.303} & \textbf{0.471} & 0.388           & 0.570          & 0.292          & 0.472          \\ \cline{2-8} 
    \multicolumn{1}{c|}{}                             & \multicolumn{1}{c|}{Enhancement}     & 1.59\%         & 2.15\%         & -37.5\%         & -35.0\%        & -1.58\%        & -1.46\%        \\ \hline
                                                      &                                      &                &                &                 &                &                &               
    \end{tabular}
    \label{tab:enhancement}
\end{table}

\subsection{Enhancing Transformers Performance}\label{sec:enhance}
We evaluated the performance improvements obtained by replacing the standard attention mechanism with XicorAttention in three state-of-the-art Transformer-based models: PatchTST, iTransformer, and TimeXer. 
Representative results for the ETTm2, Exchange, and Traffic datasets are presented in Table~\ref{tab:enhancement}, with the full results available in the Appendix~\ref{apd:sec:enhancement}. Improvements in forecasting performance are highlighted in bold. 

Experimental results show that integrating XicorAttention with PatchTST generally leads to performance improvements across the evaluated datasets, achieving a maximum improvement rate of 9.1\%. 
However, when integrated with iTransformer and TimeXer, performance gains are relatively small or sometimes even negative. 
This deterioration is particularly pronounced on the Traffic dataset, which has a large number of variables ($C=862$). 

This phenomenon can be explained by the different emphases of these models: PatchTST primarily captures temporal dependencies through patching, whereas iTransformer and TimeXer emphasize capturing inter-variable relationships. 
We provide further qualitative analysis and insights regarding this behavior in the following section (Section~\ref{subsec:qualitative}).

\section{Discussions}

\subsection{Understanding Performance Degradation in Inter-Variable Models}\label{subsec:qualitative}

\begin{figure}[t]
  \centering
  \includegraphics[width=0.48\textwidth]{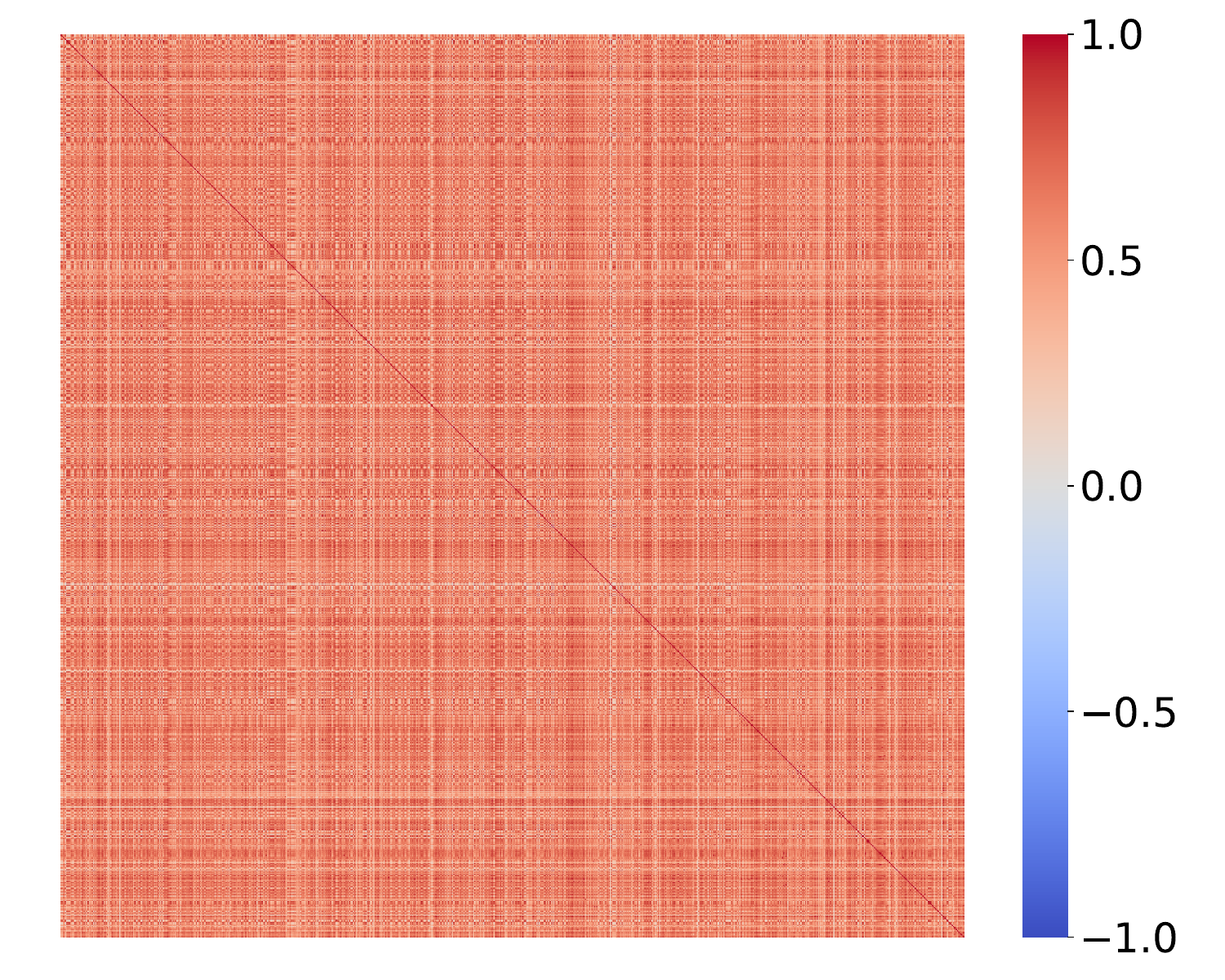}
  \caption{Heatmap illustrating pairwise correlation coefficients between variables in the Traffic dataset. Most variable pairs exhibit positive correlations.}
  \label{fig:heatmap}
\end{figure}

In this section, we qualitatively investigate why combining XicorAttention with models focusing on inter-variable relationships (e.g., iTransformer and TimeXer) sometimes leads to deteriorated performance, especially on the Weather dataset. 
To this end, we computed the pairwise correlation matrix for all variables in the Weather dataset and visualized the result in the heatmap shown in Figure~\ref{fig:heatmap}. 
Although this heatmap represents the overall correlations rather than local patterns within lookback windows or patches, it clearly indicates that most pairs of variables exhibit strong positive correlations. 
This observation suggests that relationships between these variables are predominantly linear.

Chatterjee \cite{chatterjee2021new} empirically shows a limitation of the $\xi_n$ correlation coefficient: it tends to have lower detection power when signals are smooth and non-oscillatory (e.g., linear or heteroskedastic signals). 
Thus, our qualitative analysis suggests that this limitation might have prevented XicorAttention from effectively capturing the linear inter-variable relationships, resulting in performance degradation. 

Lin et al. \cite{lin2023boosting} have recently proposed methods to overcome this limitation. 
Therefore, integrating such improvements into our approach may help mitigate these performance issues in future research.

\subsection{Hyperparameter Sensitivity}
\begin{figure}[t]
  \includegraphics[width=0.98\textwidth]{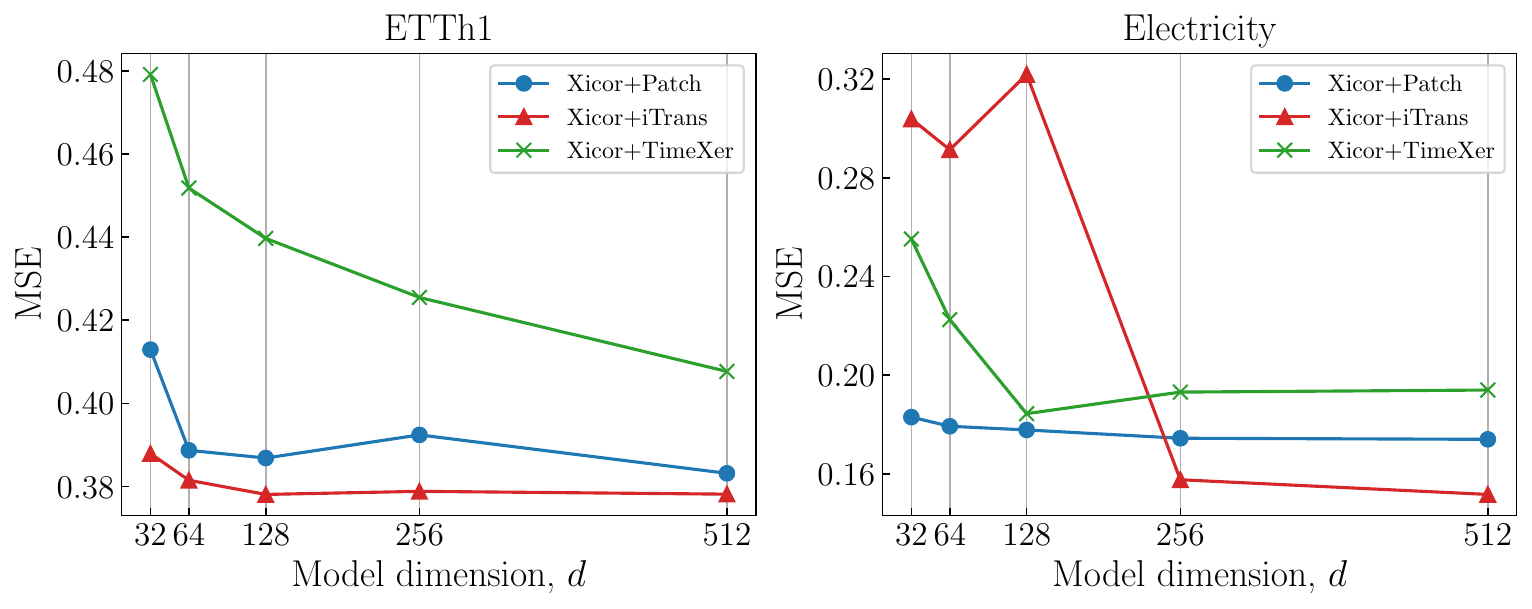}
  \caption{Sensitivity analysis of forecasting performance (MSE) with respect to model dimension $d \in \{32,64,128,256,512\}$ on ETTh1 and Electricity datasets ($H=96$). The total attention dimension $D$ is fixed at 512.} 
  \label{fig:dhead}
\end{figure}

We analyze the sensitivity of our method to the dimension $d$ of each attention head, corresponding to the sample size $n$ for estimating the coefficient $\xi_n$. 
Consider the total attention dimension $D=512$, we varied the number of attention heads $n_{\text{head}}\in\{1,2,4,8,16\}$, resulting in head dimensions $d\in\{32,64,128,256,512\}$. 
Figure~\ref{fig:dhead} shows results for the ETTh1 and Electricity datasets with forecasting horizon $H=96$.

Prediction accuracy generally improves with increasing head dimension $d$, aligning with theoretical expectations that larger sample sizes improve estimation of nonlinear dependencies. 
Performance notably deteriorates at small dimensions ($d=32$), likely due to insufficient samples. 
However, some models (e.g., ``Xicor+Patch'' on Electricity) remain robust even at smaller dimensions, suggesting dataset-specific factors. 
For practical use, we recommend a head dimension $d\geq128$, as performance stabilizes beyond this point.

\subsection{Computational Cost Analysis}
\begin{figure}[t]
  \includegraphics[width=0.98\textwidth]{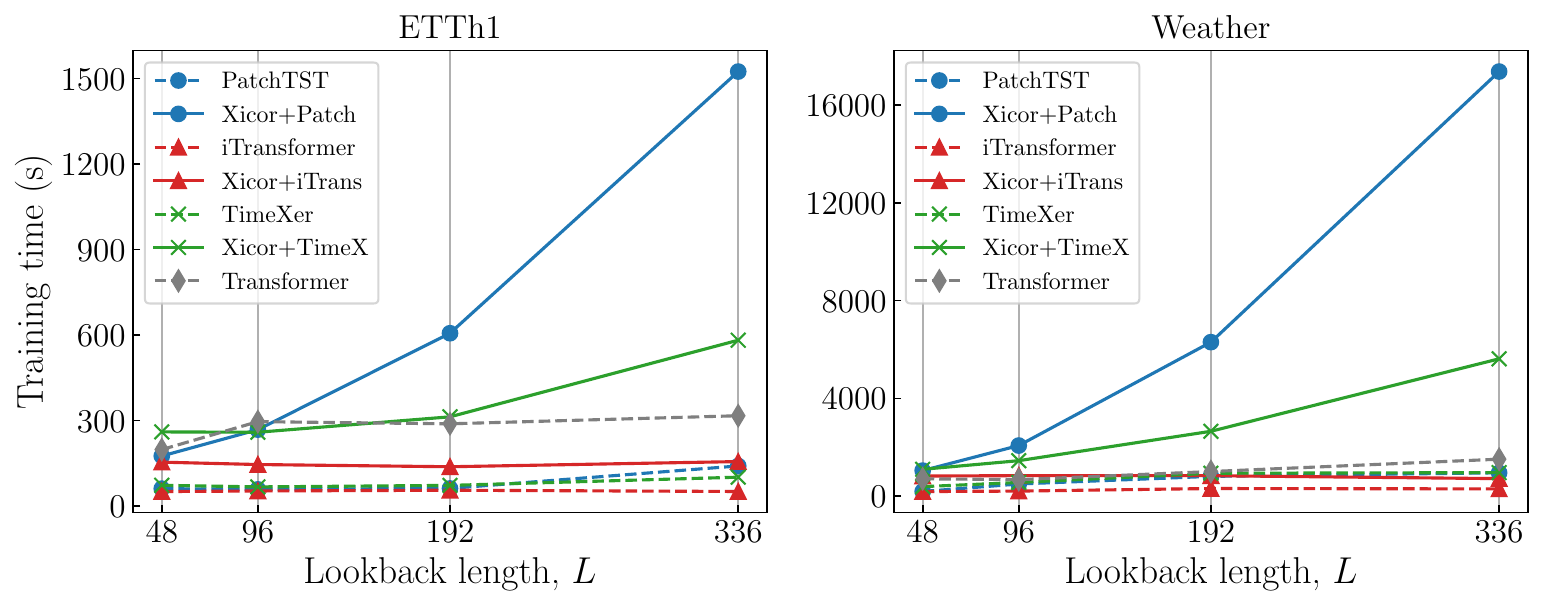}
  \caption{ Training time (in seconds) comparison across varying lookback lengths $T \in \{48, 96, 192, 336\}$ on ETTh1 and Weather datasets with a fixed forecasting horizon $H=96$.} 
  \label{fig:compute_cost}
\end{figure}

We evaluated the computational cost of XicorAttention by comparing training times for lookback lengths $T \in \{48, 96, 192, 336\}$ on the ETTh1 and Weather datasets, with forecasting horizon $H=96$, using a single NVIDIA RTX A6000 GPU (Figure~\ref{fig:compute_cost}). 
Unsurprisingly, XicorAttention incurs higher computational cost due to the sorting and ranking operations required by the $\xi_n$ coefficient. 
This overhead is particularly significant when integrated with PatchTST due to its large number of input tokens. 
However, the overhead is relatively minor with iTransformer and moderate with TimeXer. 
Future work includes optimizing the sorting and ranking operations to reduce computational cost.

\section{Conclusion and Future Work}
In this paper, we proposed XicorAttention, a novel attention mechanism designed to capture the nonlinear dependencies inherent in time series data. 
Our method computes attention weight using Chatterjee's $\xi$ correlation coefficient. 
Extensive experiments demonstrate that the proposed method achieves superior forecasting performance compared to existing Transformers on most datasets. 
In addition, integrating XicorAttention into these models can further improve their forecasting capabilities.
However, we also observed that for models emphasizing inter-variable relationships, our approach occasionally resulted in performance deterioration. A qualitative analysis suggests this behavior may stem from an inherent limitation of the $\xi$ coefficient, which struggles to detect smooth or non-oscillatory dependencies.
Therefore, integrating a modified version of the $\xi$ coefficient designed to address this limitation may further enhance forecasting accuracy, which remains a promising direction for future research. 
Additionally, since our current implementation of XicorAttention relies on sorting and ranking algorithms, another important future work is to develop computationally efficient strategies.

\bibliographystyle{splncs04}
\bibliography{references}

\newpage
\appendix
\section{Full Results}\label{apd:sec:full_result}
This appendix provides the complete experimental results omitted from the main text due to space limitations.

\subsection{Full Results of Forecasting Performance}\label{apd:sec:full_performance}
We compare the forecasting performance of XicorAttention. All results are shown in Table~\ref{apd:tab:full_result}.

\begin{sidewaystable}[p]
    \caption{Full results of Transformers with our XicorAttention. The best values are in bold, the second-best value are underlined, and the better result between original and XicorAttention are italic. \{96, 192, 336, 720\} indicates forecast horizon $H$. }    
    \resizebox{\textwidth}{!}{
        \begin{tabular}{cc|cccccccccccccccccccc}
            \hline
            \multicolumn{2}{c|}{\multirow{2}{*}{Models}}            & \multicolumn{2}{c}{\textbf{Xicor+TimeX}}                    & \multicolumn{2}{c}{TimeXer}                                 & \multicolumn{2}{c}{\textbf{Xicor+iTrans}}       & \multicolumn{2}{c}{iTransformer}                      & \multicolumn{2}{c}{\textbf{Xicor+Patch}}              & \multicolumn{2}{c}{PatchTST}                    & \multicolumn{2}{c}{FEDformer} & \multicolumn{2}{c}{Informer} & \multicolumn{2}{c}{Autoformer} & \multicolumn{2}{c}{Transformer} \\
            \multicolumn{2}{c|}{}                                   & \multicolumn{2}{c}{\textbf{(Ours)}}                         & \multicolumn{2}{c}{cite}                                    & \multicolumn{2}{c}{\textbf{(Ours)}}             & \multicolumn{2}{c}{cite}                              & \multicolumn{2}{c}{\textbf{(Ours)}}                   & \multicolumn{2}{c}{cite}                        & \multicolumn{2}{c}{cite}      & \multicolumn{2}{c}{cite}     & \multicolumn{2}{c}{cite}       & \multicolumn{2}{c}{cite}        \\ \hline
            \multicolumn{2}{c|}{Metric}                             & MAE                             & MSE                       & MAE                       & MSE                             & MAE                    & MSE                    & MAE                       & MSE                       & MAE                       & MSE                       & MAE                    & MSE                    & MAE       & MSE               & MAE           & MSE          & MAE            & MSE           & MAE            & MSE            \\ \hline
            \multicolumn{1}{c|}{\multirow{4}{*}{ETTh1}}       & 96  & \textit{\textbf{0.39744}}       & 0.384717                  & 0.402865                  & \textit{\textbf{0.38181}}       & \textit{0.40743}       & \textit{0.39383}       & 0.40979                   & 0.394974                  & 0.404009                  & 0.388893                  & {\ul \textit{0.39839}} & {\ul \textit{0.37834}} & 0.418466  & \textbf{0.37717}  & 0.771244      & 0.951099     & 0.482015       & 0.504972      & 0.741412       & 0.881161       \\
            \multicolumn{1}{c|}{}                             & 192 & {\ul \textit{\textbf{0.43152}}} & 0.439657                  & 0.435455                  & \textit{\textbf{0.42851}}       & 0.442484               & 0.452016               & \textit{0.44116}          & \textit{0.44853}          & \textit{\textbf{0.43017}} & 0.435472                  & 0.431582               & {\ul \textit{0.4249}}  & 0.444142  & \textbf{0.41998}  & 0.789581      & 1.01588      & 0.470682       & 0.476891      & 0.751061       & 0.897921       \\
            \multicolumn{1}{c|}{}                             & 336 & 0.457903                        & 0.485662                  & \textit{\textbf{0.44857}} & {\ul \textit{\textbf{0.46775}}} & 0.468993               & 0.499197               & \textit{0.46538}          & \textit{0.49213}          & {\ul \textit{0.45048}}    & 0.473156                  & 0.457633               & \textit{0.47008}       & 0.466752  & \textbf{0.45856}  & 0.781669      & 1.029791     & 0.500333       & 0.518984      & 0.809313       & 0.987317       \\
            \multicolumn{1}{c|}{}                             & 720 & 0.506333                        & 0.534783                  & \textit{\textbf{0.46123}} & {\ul \textit{\textbf{0.46953}}} & \textit{0.50076}       & \textit{0.51719}       & 0.504103                  & 0.521519                  & {\ul \textit{0.46557}}    & \textit{\textbf{0.46804}} & 0.507423               & 0.524717               & 0.503097  & 0.502238          & 0.876635      & 1.213151     & 0.508992       & 0.512151      & 0.809994       & 1.044443       \\ \hline
            \multicolumn{1}{c|}{\multirow{4}{*}{ETTh2}}       & 96  & \textit{\textbf{0.33751}}       & {\ul 0.286291}            & {\ul 0.337737}            & \textit{\textbf{0.28596}}       & \textit{0.34733}       & \textit{0.29396}       & 0.349618                  & 0.300403                  & 0.34731                   & 0.299818                  & \textit{0.34593}       & \textit{0.29138}       & 0.391798  & 0.350731          & 1.335586      & 2.860698     & 0.409597       & 0.376178      & 1.313012       & 2.712585       \\
            \multicolumn{1}{c|}{}                             & 192 & {\ul 0.391971}                  & {\ul 0.374015}            & \textit{\textbf{0.39026}} & \textit{\textbf{0.36451}}       & 0.39987                & \textit{0.38089}       & \textit{0.39982}          & 0.381812                  & \textit{0.39457}          & 0.379045                  & 0.40367                & \textit{0.37807}       & 0.449716  & 0.441895          & 2.073156      & 6.124638     & 0.448578       & 0.443318      & 1.974933       & 6.049862       \\
            \multicolumn{1}{c|}{}                             & 336 & 0.441056                        & 0.464549                  & \textit{\textbf{0.42329}} & \textit{\textbf{0.41089}}       & 0.435531               & 0.42756                & \textit{0.43235}          & \textit{0.42355}          & {\ul \textit{0.42663}}    & {\ul \textit{0.41696}}    & 0.440279               & 0.424984               & 0.490607  & 0.498713          & 1.944658      & 5.352784     & 0.479468       & 0.471696      & 1.966973       & 5.918828       \\
            \multicolumn{1}{c|}{}                             & 720 & 0.464895                        & 0.428683                  & \textit{\textbf{0.43209}} & \textit{\textbf{0.407}}         & 0.44965                & 0.434097               & \textit{0.44513}          & \textit{0.42642}          & {\ul \textit{0.43921}}    & {\ul \textit{0.42011}}    & 0.453828               & 0.435533               & 0.487303  & 0.480462          & 1.726181      & 4.285115     & 0.491792       & 0.484533      & 1.456909       & 3.219348       \\ \hline
            \multicolumn{1}{c|}{\multirow{4}{*}{ETTm1}}       & 96  & {\ul 0.359269}                  & {\ul 0.322705}            & \textit{\textbf{0.35632}} & \textit{\textbf{0.31783}}       & \textit{0.37223}       & 0.341643               & 0.376444                  & \textit{0.34128}          & 0.365382                  & 0.324542                  & \textit{0.36445}       & \textit{0.32357}       & 0.411741  & 0.365792          & 0.555926      & 0.61913      & 0.494043       & 0.541412      & 0.650005       & 0.770603       \\
            \multicolumn{1}{c|}{}                             & 192 & {\ul 0.384243}                  & 0.367569                  & \textit{\textbf{0.38295}} & \textit{\textbf{0.36158}}       & \textit{0.39278}       & 0.384725               & 0.394902                  & \textit{0.38062}          & \textit{0.38689}          & {\ul \textit{0.3644}}     & 0.391785               & 0.372005               & 0.446818  & 0.435941          & 0.618007      & 0.729173     & 0.511225       & 0.566433      & 0.66784        & 0.797684       \\
            \multicolumn{1}{c|}{}                             & 336 & {\ul 0.406968}                  & 0.399197                  & \textit{\textbf{0.4067}}  & \textit{\textbf{0.39512}}       & \textit{0.41582}       & 0.420613               & 0.41892                   & \textit{0.41908}          & 0.408516                  & {\ul \textit{0.39513}}    & \textit{0.40775}       & 0.39901                & 0.475748  & 0.483518          & 0.893438      & 1.250256     & 0.526109       & 0.595881      & 0.821441       & 1.093508       \\
            \multicolumn{1}{c|}{}                             & 720 & \textit{\textbf{0.44085}}       & 0.456164                  & {\ul 0.441279}            & {\ul \textit{\textbf{0.45235}}} & \textit{0.45498}       & 0.48944                & 0.455596                  & \textit{0.48628}          & \textit{0.44147}          & \textit{\textbf{0.45076}} & 0.444542               & 0.458366               & 0.48861   & 0.506704          & 0.731915      & 0.950557     & 0.520485       & 0.56551       & 0.826463       & 1.150679       \\ \hline
            \multicolumn{1}{c|}{\multirow{4}{*}{ETTm2}}       & 96  & \textit{\textbf{0.25469}}       & \textit{\textbf{0.17081}} & {\ul 0.256219}            & {\ul 0.171342}                  & \textit{0.25825}       & \textit{0.17622}       & 0.266864                  & 0.183787                  & \textit{0.26134}          & \textit{0.17746}          & 0.267816               & 0.18552                & 0.281227  & 0.191732          & 0.461423      & 0.376779     & 0.316087       & 0.229946      & 0.529988       & 0.520615       \\
            \multicolumn{1}{c|}{}                             & 192 & {\ul 0.299866}                  & {\ul 0.237134}            & \textit{\textbf{0.29905}} & \textit{\textbf{0.23678}}       & \textit{0.30478}       & \textit{0.24359}       & 0.312499                  & 0.252822                  & \textit{0.30326}          & \textit{0.24222}          & 0.310155               & 0.250452               & 0.325829  & 0.263905          & 0.722208      & 0.846208     & 0.332241       & 0.273994      & 0.78116        & 1.085032       \\
            \multicolumn{1}{c|}{}                             & 336 & 0.341301                        & {\ul 0.301199}            & \textit{\textbf{0.33883}} & \textit{\textbf{0.29732}}       & \textit{0.34871}       & \textit{0.30984}       & 0.351524                  & 0.314952                  & {\ul \textit{0.3407}}     & \textit{0.3013}           & 0.349418               & 0.311905               & 0.367466  & 0.330658          & 0.900632      & 1.376828     & 0.373431       & 0.342348      & 0.784564       & 1.084447       \\
            \multicolumn{1}{c|}{}                             & 720 & \textit{\textbf{0.39449}}       & \textit{\textbf{0.39541}} & 0.397398                  & {\ul 0.39774}                   & \textit{0.40174}       & \textit{0.40652}       & 0.406011                  & 0.411867                  & {\ul \textit{0.39654}}    & \textit{0.39809}          & 0.415177               & 0.423176               & 0.422897  & 0.42609           & 1.541553      & 4.298708     & 0.424243       & 0.43111       & 1.18187        & 2.508886       \\ \hline
            \multicolumn{1}{c|}{\multirow{4}{*}{Exchange}}    & 96  & \textit{\textbf{0.20829}}       & \textit{\textbf{0.08893}} & 0.216696                  & 0.09489                         & 0.208385               & 0.087758               & {\ul \textit{0.20715}}    & {\ul \textit{0.08699}}    & \textit{\textbf{0.20178}} & \textit{\textbf{0.08347}} & 0.215837               & 0.097362               & 0.294307  & 0.166371          & 0.778332      & 0.93537      & 0.279267       & 0.148764      & 0.567765       & 0.534821       \\
            \multicolumn{1}{c|}{}                             & 192 & \textit{0.30428}                & \textit{0.18359}          & 0.3048                    & 0.184451                        & 0.305976               & 0.182972               & {\ul \textit{0.30319}}    & \textit{\textbf{0.17994}} & \textit{\textbf{0.30276}} & {\ul \textit{0.18102}}    & 0.303602               & 0.182193               & 0.384393  & 0.27863           & 0.84333       & 1.116397     & 0.389078       & 0.285756      & 0.741638       & 0.965695       \\
            \multicolumn{1}{c|}{}                             & 336 & 0.427647                        & 0.350847                  & {\ul \textit{0.41574}}    & \textit{0.33136}                & \textit{0.41738}       & {\ul \textit{0.33059}} & 0.419631                  & 0.334185                  & \textit{\textbf{0.40573}} & \textit{\textbf{0.3135}}  & 0.426069               & 0.342374               & 0.504778  & 0.472605          & 0.9849        & 1.503983     & 0.519782       & 0.485186      & 0.943076       & 1.462384       \\
            \multicolumn{1}{c|}{}                             & 720 & \textit{0.72096}                & \textit{0.92236}          & 0.767678                  & 1.027287                        & {\ul \textit{0.69645}} & {\ul \textit{0.854}}   & 0.700377                  & 0.856194                  & \textit{\textbf{0.68563}} & \textit{\textbf{0.83145}} & 0.731269               & 0.950503               & 0.828927  & 1.166216          & 1.414329      & 2.931928     & 0.820404       & 1.112792      & 1.325629       & 2.552118       \\ \hline
            \multicolumn{1}{c|}{\multirow{4}{*}{Weather}}     & 96  & \textit{\textbf{0.20018}}       & \textit{\textbf{0.15419}} & {\ul 0.204677}            & {\ul 0.157357}                  & \textit{0.21484}       & \textit{0.1753}        & 0.215906                  & 0.175339                  & \textit{0.21417}          & \textit{0.1733}           & 0.21802                & 0.175079               & 0.300086  & 0.220547          & 0.412407      & 0.350854     & 0.371566       & 0.333764      & 0.443417       & 0.428663       \\
            \multicolumn{1}{c|}{}                             & 192 & {\ul 0.249669}                  & {\ul 0.205953}            & \textit{\textbf{0.24786}} & \textit{\textbf{0.20429}}       & \textit{0.25766}       & \textit{0.22429}       & 0.257849                  & 0.225247                  & \textit{0.25422}          & \textit{0.2193}           & 0.256266               & 0.221199               & 0.344125  & 0.278168          & 0.587349      & 0.680014     & 0.373423       & 0.314936      & 0.551271       & 0.602256       \\
            \multicolumn{1}{c|}{}                             & 336 & 0.295495                        & {\ul 0.268112}            & \textit{\textbf{0.29067}} & \textit{\textbf{0.26209}}       & \textit{0.29786}       & 0.280761               & 0.298146                  & \textit{0.27967}          & {\ul \textit{0.29425}}    & \textit{0.27345}          & 0.298165               & 0.279941               & 0.381841  & 0.339278          & 0.443778      & 0.426998     & 0.387616       & 0.352182      & 0.602792       & 0.672673       \\
            \multicolumn{1}{c|}{}                             & 720 & 0.345915                        & {\ul 0.345821}            & \textit{\textbf{0.34012}} & \textit{\textbf{0.33979}}       & \textit{0.3488}        & \textit{0.35896}       & 0.350979                  & 0.361121                  & {\ul \textit{0.34458}}    & \textit{0.35049}          & 0.348518               & 0.356482               & 0.418015  & 0.409561          & 0.760245      & 1.056531     & 0.521782       & 0.557259      & 0.696712       & 0.867163       \\ \hline
            \multicolumn{1}{c|}{\multirow{4}{*}{Electricity}} & 96  & 0.260066                        & 0.158127                  & {\ul \textit{0.24234}}    & \textit{\textbf{0.14042}}       & 0.245635               & 0.152425               & \textit{\textbf{0.24019}} & {\ul \textit{0.14857}}    & \textit{0.26225}          & \textit{0.1698}           & 0.273041               & 0.180144               & 0.309201  & 0.195007          & 0.414652      & 0.33128      & 0.315183       & 0.199383      & 0.411694       & 0.316787       \\
            \multicolumn{1}{c|}{}                             & 192 & 0.262977                        & 0.164516                  & {\ul \textit{0.25602}}    & \textit{\textbf{0.15751}}       & 0.263906               & 0.17091                & \textit{\textbf{0.25559}} & {\ul \textit{0.16423}}    & \textit{0.27589}          & \textit{0.18168}          & 0.279611               & 0.187397               & 0.31543   & 0.202173          & 0.436655      & 0.355623     & 0.358076       & 0.276127      & 0.439515       & 0.36181        \\
            \multicolumn{1}{c|}{}                             & 336 & 0.280103                        & 0.181579                  & {\ul \textit{0.27303}}    & \textit{\textbf{0.17439}}       & 0.275703               & 0.182875               & \textit{\textbf{0.27163}} & {\ul \textit{0.1791}}     & \textit{0.28903}          & \textit{0.19709}          & 0.295867               & 0.204222               & 0.34254   & 0.229432          & 0.434061      & 0.351395     & 0.353328       & 0.243466      & 0.49321        & 0.442858       \\
            \multicolumn{1}{c|}{}                             & 720 & \textit{\textbf{0.30678}}       & \textit{\textbf{0.21076}} & {\ul 0.307849}            & {\ul 0.211396}                  & \textit{0.31145}       & \textit{0.22201}       & 0.312379                  & 0.227937                  & \textit{0.3225}           & \textit{0.23763}          & 0.328327               & 0.245563               & 0.366381  & 0.263329          & 0.460772      & 0.401391     & 0.386181       & 0.307746      & 0.52677        & 0.492606       \\ \hline
            \multicolumn{1}{c|}{\multirow{4}{*}{Traffic}}     & 96  & 0.274738                        & 0.437136                  & {\ul \textit{0.27171}}    & {\ul \textit{0.42922}}          & 0.377682               & 0.541858               & \textit{\textbf{0.26868}} & \textit{\textbf{0.39316}} & \textit{0.29159}          & \textit{0.44718}          & 0.298355               & 0.458686               & 0.362563  & 0.579718          & 0.412484      & 0.73532      & 0.397315       & 0.662156      & 0.363599       & 0.653978       \\
            \multicolumn{1}{c|}{}                             & 192 & 0.288482                        & 0.459019                  & {\ul \textit{0.28194}}    & {\ul \textit{0.44875}}          & 0.387678               & 0.566403               & \textit{\textbf{0.277}}   & \textit{\textbf{0.41278}} & \textit{0.29364}          & \textit{0.45813}          & 0.300787               & 0.468703               & 0.378525  & 0.605474          & 0.432518      & 0.765568     & 0.417553       & 0.658241      & 0.362597       & 0.658864       \\
            \multicolumn{1}{c|}{}                             & 336 & 0.292311                        & 0.477308                  & {\ul \textit{0.28901}}    & {\ul \textit{0.47012}}          & 0.381303               & 0.5625                 & \textit{\textbf{0.28304}} & \textit{\textbf{0.42492}} & \textit{0.30647}          & \textit{0.47491}          & 0.307436               & 0.483028               & 0.380606  & 0.613424          & 0.499816      & 0.886805     & 0.401511       & 0.636649      & 0.362532       & 0.662322       \\
            \multicolumn{1}{c|}{}                             & 720 & 0.312468                        & 0.517199                  & {\ul \textit{0.30709}}    & \textit{0.51616}                & 0.406453               & 0.61044                & \textit{\textbf{0.30098}} & \textit{\textbf{0.45966}} & \textit{0.32175}          & {\ul \textit{0.50676}}    & 0.326231               & 0.517871               & 0.394746  & 0.64132           & 0.574479      & 1.011161     & 0.431573       & 0.689229      & 0.376272       & 0.695434       \\ \hline
            \end{tabular}            
        }
    \label{apd:tab:full_result}
\end{sidewaystable}

\subsection{Full Results of Enhancing Transformers Performance}\label{apd:sec:enhancement}
We evaluated the performance improvements obtained by replacing the standard attention mechanism with XicorAttention in three state-of-the-art Transformer-based models: PatchTST, iTransformer, and TimeXer. 
All enhancement results are presented in Table~\ref{apd:full_enh}.
\begin{table}[htbp]
    \caption{Full Results of performance enhancement (\%) by replacing original attention with XicorAttention (+Xicor) in PatchTST, iTransformer, and TimeXer. Bold indicates the better result between original and XicorAttention.}
    \centering
    \begin{tabular}{cccccccc}
        \hline
        \multicolumn{2}{c|}{Models}                                                              & \multicolumn{2}{c}{PatchTST}    & \multicolumn{2}{c}{iTransformer} & \multicolumn{2}{c}{TimeXer}     \\ \hline
        \multicolumn{2}{c|}{Metric}                                                              & MAE            & MSE            & MAE             & MSE            & MAE            & MSE            \\ \hline
        \multicolumn{1}{c|}{\multirow{3}{*}{ETTh1}}       & \multicolumn{1}{c|}{Original}        & 0.447          & 0.446          & 0.454           & \textbf{0.462} & \textbf{0.436} & \textbf{0.435} \\
        \multicolumn{1}{c|}{}                             & \multicolumn{1}{c|}{\textbf{+Xicor}} & \textbf{0.437} & \textbf{0.440} & \textbf{0.454}  & 0.463          & 0.447          & 0.458          \\ \cline{2-8} 
        \multicolumn{1}{c|}{}                             & \multicolumn{1}{c|}{Enhancement}     & 2.25\%        & 1.38\%        & 0.04\%         & -0.27\%       & -2.31\%       & -5.15\%       \\ \hline
        \multicolumn{1}{c|}{\multirow{3}{*}{ETTh2}}       & \multicolumn{1}{c|}{Original}        & 0.409          & 0.378          & \textbf{0.405}  & \textbf{0.379} & \textbf{0.394} & \textbf{0.363} \\
        \multicolumn{1}{c|}{}                             & \multicolumn{1}{c|}{\textbf{+Xicor}} & \textbf{0.400} & \textbf{0.376} & 0.406           & 0.380          & 0.406          & 0.382          \\ \cline{2-8} 
        \multicolumn{1}{c|}{}                             & \multicolumn{1}{c|}{Enhancement}     & 2.05\%        & 0.60\%        & -0.27\%        & -0.08\%       & -2.99\%       & -5.17\%       \\ \hline
        \multicolumn{1}{c|}{\multirow{3}{*}{ETTm1}}       & \multicolumn{1}{c|}{Original}        & 0.401          & 0.385          & 0.410           & \textbf{0.403} & \textbf{0.396} & \textbf{0.379} \\
        \multicolumn{1}{c|}{}                             & \multicolumn{1}{c|}{\textbf{+Xicor}} & \textbf{0.400} & \textbf{0.381} & \textbf{0.408}  & 0.406          & 0.397          & 0.383          \\ \cline{2-8} 
        \multicolumn{1}{c|}{}                             & \multicolumn{1}{c|}{Enhancement}     & 0.38\%         & 1.10\%         & 0.63\%          & -0.55\%        & -0.28\%        & -1.27\%        \\ \hline
        \multicolumn{1}{c|}{\multirow{3}{*}{ETTm2}}       & \multicolumn{1}{c|}{Original}        & 0.331          & 0.280          & 0.330           & 0.279          & 0.319          & \textbf{0.263} \\
        \multicolumn{1}{c|}{}                             & \multicolumn{1}{c|}{\textbf{+Xicor}} & \textbf{0.322} & \textbf{0.268} & \textbf{0.324}  & \textbf{0.271} & \textbf{0.318} & 0.264          \\ \cline{2-8} 
        \multicolumn{1}{c|}{}                             & \multicolumn{1}{c|}{Enhancement}     & 2.91\%         & 4.24\%         & 1.89\%          & 2.68\%         & 0.08\%         & -0.14\%        \\ \hline
        \multicolumn{1}{c|}{\multirow{3}{*}{Exchange}}    & \multicolumn{1}{c|}{Original}        & 0.378          & 0.276          & \textbf{0.369}  & \textbf{0.259} & 0.381          & 0.278          \\
        \multicolumn{1}{c|}{}                             & \multicolumn{1}{c|}{\textbf{+Xicor}} & \textbf{0.361} & \textbf{0.251} & 0.369           & 0.259          & \textbf{0.374} & \textbf{0.270} \\ \cline{2-8} 
        \multicolumn{1}{c|}{}                             & \multicolumn{1}{c|}{Enhancement}     & 4.48\%        & 9.12\%         & -0.10\%         & -0.30\%        & 1.88\%         & 2.96\%         \\ \hline
        \multicolumn{1}{c|}{\multirow{3}{*}{Weather}}     & \multicolumn{1}{c|}{Original}        & 0.276          & 0.249          & 0.276           & 0.251          & \textbf{0.266}          & \textbf{0.231}          \\
        \multicolumn{1}{c|}{}                             & \multicolumn{1}{c|}{\textbf{+Xicor}} & \textbf{0.273} & \textbf{0.246} & \textbf{0.275}  & \textbf{0.251} & 0.267 & 0.233 \\ \cline{2-8} 
        \multicolumn{1}{c|}{}                             & \multicolumn{1}{c|}{Enhancement}     & 1.25\%         & 1.47\%         & 0.32\%          & 0.17\%         & -0.46\%        & -0.71\%        \\ \hline
        \multicolumn{1}{c|}{\multirow{3}{*}{Electricity}} & \multicolumn{1}{c|}{Original}        & 0.293          & 0.203          & \textbf{0.269}  & \textbf{0.178} & \textbf{0.269} & \textbf{0.169} \\
        \multicolumn{1}{c|}{}                             & \multicolumn{1}{c|}{\textbf{+Xicor}} & \textbf{0.287} & \textbf{0.195} & 0.273           & 0.180          & 0.277          & 0.178          \\ \cline{2-8} 
        \multicolumn{1}{c|}{}                             & \multicolumn{1}{c|}{Enhancement}     & 2.35\%         & 3.88\%         & -1.67\%         & -1.51\%        & -3.03\%        & -5.12\%        \\ \hline
        \multicolumn{1}{c|}{\multirow{3}{*}{Traffic}}     & \multicolumn{1}{c|}{Original}        & 0.308          & 0.482          & \textbf{0.282}  & \textbf{0.422} & \textbf{0.287} & \textbf{0.465} \\
        \multicolumn{1}{c|}{}                             & \multicolumn{1}{c|}{\textbf{+Xicor}} & \textbf{0.303} & \textbf{0.471} & 0.388           & 0.570          & 0.292          & 0.472          \\ \cline{2-8} 
        \multicolumn{1}{c|}{}                             & \multicolumn{1}{c|}{Enhancement}     & 1.59\%         & 2.15\%         & -37.5\%         & -35.0\%        & -1.58\%        & -1.46\%        \\ \hline
                                                          &                                      &                &                &                 &                &                &               
        \end{tabular}
        \label{apd:full_enh}
  \end{table}

\end{document}